\documentclass[11pt]{article} 
\usepackage{fullpage}
\usepackage{times}  
\usepackage{helvet}  
\usepackage{courier}  
\usepackage[hyphens]{url}  
\usepackage{graphicx} 
\usepackage{natbib}
\usepackage{caption}
\usepackage[most]{tcolorbox}
\usepackage{xcolor}
\usepackage{amssymb}
\usepackage{booktabs}
\usepackage{pifont}
\usepackage{bm}
\usepackage{hyperref}
\usepackage{booktabs}
\usepackage{array}
\usepackage{enumitem}
\newcommand{\tabitem}{~~\llap{\textbullet}~~}
\tcbuselibrary{listingsutf8}

\usepackage{algorithm}
\usepackage{algorithmic}

\usepackage{newfloat}
\usepackage{listings}
\DeclareCaptionStyle{ruled}{labelfont=normalfont,labelsep=colon,strut=off} 
\floatstyle{ruled}
\newfloat{listing}{tb}{lst}{}
\floatname{listing}{Listing}

\title{\textit{TalkToAgent}: A Human-centric Explanation of Reinforcement Learning Agents with Large Language Models}

\author{
  Haechang Kim\thanks{Department of Chemical and Biological Engineering, Seoul National University, Republic of Korea}
  \and
  Hao Chen\thanks{Davidson School of Chemical Engineering, Purdue University, United States}
  \and
  Can Li\footnotemark[2]
  \and
  Jong Min Lee\footnotemark[1]
}

\date{}

\begin{document}

\maketitle

\begin{abstract}
Explainable Reinforcement Learning (XRL) has emerged as a promising approach in improving the transparency of Reinforcement Learning (RL) agents. However, there remains a gap between complex RL policies and domain experts, due to the limited comprehensibility of XRL results and isolated coverage of current XRL approaches that leave users uncertain about which tools to employ. To address these challenges, we introduce TalkToAgent, a multi-agent Large Language Models (LLM) framework that delivers interactive, natural language explanations for RL policies. The architecture with five specialized LLM agents--\textit{Coordinator, Explainer, Coder, Evaluator, and Debugger}-- enables TalkToAgent to automatically map user queries to relevant XRL tools and clarify an agent's actions in terms of either key state variables, expected outcomes, or counterfactual explanations. Moreover, our approach extends previous counterfactual explanations by deriving alternative scenarios from qualitative behavioral descriptions, or even new rule-based policies. We validated TalkToAgent on quadruple-tank process control problem, a well-known nonlinear control benchmark. Results demonstrated that TalkToAgent successfully mapped user queries into XRL tasks with high accuracy, and coder–debugger interactions minimized failures in counterfactual generation. Furthermore, qualitative evaluation confirmed that TalkToAgent effectively interpreted agent's actions and contextualized their meaning within the problem domain.
\end{abstract}

\section{Introduction}
\label{introduction}

Deep reinforcement learning (DRL) has significantly improved the scalability of RL algorithms by leveraging deep neural networks to handle high-dimensional state and action spaces. This advancement has enabled broader adoption of RL, including in safety-critical industrial applications such as process control or energy systems \citep{park2025reinforcement, perera2021applications}. Despite these advancements, the inherent black-box nature of DRL models makes their decision-making difficult to interpret. This is particularly critical in safety-sensitive domains, where trust and transparency are essential. While explainable reinforcement learning (XRL) has demonstrated promise in interpreting the agent's behavior \citep{milani2024explainable, puiutta2020explainable}, there still exist critical challenges that hinder the direct adoption of existing XRL methods in industrial applications.

One major challenge in current XRL approaches is that most methods offer only a limited view of the agent’s behavior. Since RL agents make sequential decisions aimed at maximizing future rewards, a wider range of factors influence the agent's actions compared to supervised learning settings. For example, an action may depend not only on the current state, but also on its long-term consequences. To capture this broader perspective, an effective XRL framework should flexibly support diverse explanation types aligned with user intent.

In addition, the lack of familiarity with XRL techniques makes it challenging for non-experts to effectively utilize them. Users may struggle to identify which XRL method is most appropriate for addressing their question. Even when relevant XRL outputs are provided, deriving meaningful interpretations from them is still nontrivial. This highlights the need for a natural language interface that can both map user intent to appropriate XRL strategies and deliver results in a comprehensible form.

Lastly, counterfactual explanations \citep{gajcin2024redefining}, which explain an agent's decisions by contrasting them with alternative ones, are typically limited to simple action substitutions in existing XRL studies. This hinders their practical applications where users often express intent through abstract or qualitative behavioral descriptions. For example, when users request alternative scenarios where agent's behaviors are described in natural language (e.g., \textit{conservative} or \textit{opposite} behavior), or when they want to compare the agent with new counterfactual policies, existing methods fail to accommodate such cases. Addressing these needs would significantly broaden the scope of explainability in practice.

To address these challenges, we propose \textit{TalkToAgent}, a multi-agent Large Language Model (LLM) framework that interprets RL policies through natural language interactions. \textit{TalkToAgent} is designed to handle diverse user queries, including questions about feature importance, expected outcome, and counterfactual scenarios, by mapping them to relevant XRL tools. The output from  \textit{TalkToAgent} includes figures generated from XRL tools and natural language interpretations generated by LLM agents, helping the user understand the behavior of the agent in multimodal fashion.

Another key novelty of \textit{TalkToAgent} lies in its versatile generation of counterfactual scenarios. Beyond the action-level interventions, \textit{TalkToAgent} supports comparison of diverse alternative strategies by translating high-level user requests into action trajectories or executable policy code.
Our main contributions are summarized below:

\begin{itemize}

\item We introduce \textit{TalkToAgent}, a multi-agent LLM-based XRL explainer framework that answers natural language queries about the intentions behind an agent's actions.

\item \textit{TalkToAgent} unifies solutions to various XRL tasks within a single framework, achieving a mapping accuracy of 96.7\% with the GPT-4.1 model.

\item We expand counterfactual explanations beyond single-step actions to include qualitative behavioral terms and rule-based policies, while attaining minimal failure rate through interactions between coder and evaluator agents.

\item \textit{TalkToAgent} improves the interpretability of RL agents by providing domain-aware textual explanations of XRL visualizations.

\end{itemize}

The source code is available at \url{https://github.com/li-group/TalktoAgent}.

\section{Related Work}
\label{Related Work}

\begin{table}[t]
\centering
\setlength{\tabcolsep}{1mm}
\begin{tabular}{p{3.2cm}cccccc}
\toprule
Previous studies & FI & EO & CF-A & CF-B & CF-P & NL \\
\midrule
\citet{xi2025model} & \ding{51} & \ding{55} & \ding{55} & \ding{55} & \ding{55} & \ding{51} \\
\citet{amitai2024explaining} & \ding{55} & \ding{51} & \ding{51} & \ding{55} & \ding{55} & \ding{55} \\
\citet{towers2024explaining} & \ding{51} & \ding{51} & \ding{51} & \ding{55} & \ding{55} & \ding{55} \\
Our work  & \ding{51} & \ding{51} & \ding{51} & \ding{51} & \ding{51} & \ding{51} \\
\bottomrule
\end{tabular}

\caption{Comparison of supported explanation types by our method and previous studies. (FI: Feature Importance, EO: Expected Outcomes, CF-A: Action-based Counterfactual, CF-B: Behavior-based Counterfactual, CF-P: Policy-based Counterfactual, NL: Natural Language)}
\label{tab:relatedwork}
\end{table}

\paragraph{Explainable Reinforcement Learning (XRL)}
XRL has emerged as a pivotal tool for enhancing the transparency and trustworthiness of RL agents, particularly in safety-sensitive domains such as autonomous driving \citep{atakishiyev2024explainable} and process control \citep{szatmari2024resilience}. Recent XRL literature can be broadly categorized into four major directions: \textit{feature importance}, \textit{expected outcome}, \textit{counterfactual}, and \textit{policy-level} explanations.

\textit{Feature importance} methods elucidate which state features or pixels most influence the agent's decisions. Traditional explainable AI techniques such as saliency maps \citep{wang2019verbal, atrey2019exploratory}, interpretable surrogate models \citep{liu2018toward}, and post-hoc attributions \citep{zhang2021explainable} have been adopted to identify critical features.

Agent's \textit{expected outcome} helps identify when and where the agent expects important events and rewards to occur from executing a certain action. It can be attained by decomposing Q-values into temporal dimension \citep{ruggeri2025explainable, towers2024explaining}, or extracting future trajectories via belief maps and forward simulations \citep{yau2020did, van2018contrastive}.

A \textit{counterfactual explanation} illustrates what would have happened if the agent had behaved differently, helping users better understand the rationale behind its actual decision \citep{miller2019explanation, byrne2019counterfactuals}. Prior works have integrated such counterfactuals with other explainable components, including future trajectories \citep{amitai2024explaining}, causal graphs \citep{madumal2020explainable}, and decomposed rewards \citep{lin2020contrastive}. These approaches typically intervene on action values within a single timestep, limiting their expressiveness in the explanations.

Lastly, \textit{policy-level} explanations offer a global perspective on the agent's behavior across the entire state space. These approaches include identifying regions where the agent behaves suboptimally or exhibits high uncertainty \citep{sequeira2020interestingness, huang2018establishing}, or analyzing the dynamics between physically meaningful clusters of states \citep{zahavy2016graying}.

Such diverse nature of XRL techniques highlights the need for systems that capture the intent behind user queries and automatically map them to the most appropriate XRL techniques. In addition, extending counterfactual explanations beyond simple action interventions to include high-level behavioral terms or alternative policies would enrich their applicability.

\paragraph{Natural Language-based Explainers}
Natural language has been gaining increasing attention in the field of explainable artificial intelligence as it provides an interactive and accessible medium for users. In supervised settings, \citet{slack2023explaining} mapped user intent to relevant explanation types such as feature importance, counterfactuals, and feature interactions. In RL settings, \citet{hayes2017improving} proposed a template-based system to answer natural language inquiries about robot behavior. \citet{xi2025model} leveraged LLMs to explain local decision paths of states influencing an agent's actions.

Building on these efforts, this work maps natural language queries directly to multiple XRL tools, and provides textual interpretations along with the resulting visualizations.

\paragraph{Relations to previous works}
We compare the types of explanations supported by our proposed TalkToAgent framework with those of related work in Table~\ref{tab:relatedwork}. The explanation types range from feature importance (FI) and expected outcomes (EO) to various forms of counterfactuals (CF-A, CF-B, CF-P). Detailed descriptions of each explanation type are provided in the Methods section.

To the best of our knowledge, the only prior work that addressed natural language-based XRL queries is \citet{xi2025model}, which focused solely on \textit{feature importance} explanations. Although \citet{amitai2024explaining} and \citet{towers2024explaining} explored counterfactual explanations by comparing future rewards, their methods are limited to single-step interventions and lack adaptability to more complex scenarios.

\section{Methods}
\label{methods}

\subsection{Overview}
\label{overview}

\begin{figure}[t]
\centering
\includegraphics[width=0.7\columnwidth]{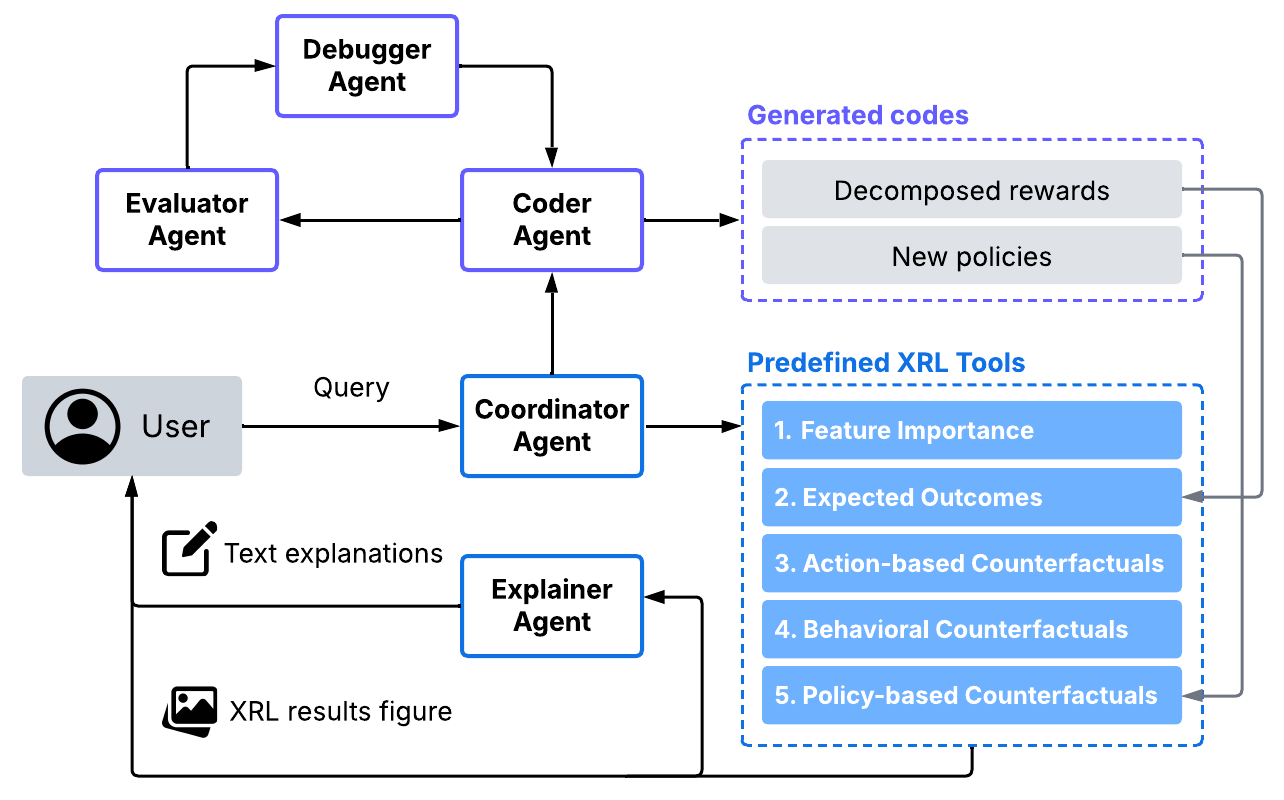} 
\caption{Overall framework of \textit{TalkToAgent}.}
\label{fig:Overall}
\end{figure}

The overall structure of \textit{TalkToAgent}, illustrated in Figure \ref{fig:Overall}, consists of five specialized LLM agents with key Python functions that implement XRL techniques.

The workflow begins when a user poses a natural language query about a trained RL agent. A set of agents collaborate to interpret the query, invoke appropriate XRL tools, and return multimodal explanations. The system also features automated code generation for new reward functions or policies, enabling flexible responses to diverse user queries. Details about each agent's specific role and their interactions are described in subsequent sections.

\subsection{Multi-agent LLM structure}
We design a multi-agent framework consisting of five specialized LLM agents, where each is assigned a distinct role and corresponding instruction in the explanation process. Through interactions among these agents, the system is able to automate the explanation pipeline.

\begin{itemize}

\item The \textbf{coordinator} interprets user queries and determines the appropriate XRL tools along with its relevant arguments. If a query requested an operation outside the available XRL function scope, the coordinator would explicitly return an error.

\item The \textbf{explainer} provides natural language explanations for the generated XRL output figures. Given domain knowledge prior to the explanation, it is further instructed to relate the interpretations to the context of the current system, offering implications for domain experts.

\item The \textbf{coder} constructs Python code when requested by the XRL function tools. In our work, the coder agent is demanded to generate two types of codes: decomposed reward functions and counterfactual rule-based policies, based on instructions from the coordinator. To reduce potential errors, it is instructed to produce a code that closely follows the structure of the original reward or policy code. The generated code serves as the user-defined rewards or policy component for assisting explanations.

\item The \textbf{evaluator} validates whether the code generated by the coder agent correctly implements the requested reward or policy. It prevents against generating hallucinated policies by detecting deviations from the intended logic and raising exceptions when necessary.

\item Whenever execution errors or hallucinations are detected by the evaluator, the \textbf{debugger} analyzes both the code and the resulting traceback to generate debugging instructions. The instructions are then sent back to the coder for iterative refinement, ensuring that the final policy is both executable and aligned with the user’s intent.

\end{itemize}

All agents are implemented using OpenAI’s GPT-4.1 model by default, with different GPT variants used for sensitivity analysis. System descriptions and environment parameters are provided as contextual prompts to ensure domain-aware reasoning. Full prompts for each agent are available in the supplementary materials.

\subsection{Automated code generation}
The code generation procedure using the agents defined previously is outlined in Algorithm \ref{alg:coder}. If the generated code fails during execution or the evaluator agent determines that the resulting trajectory does not align with the user’s intent, an error is raised. This triggers the debugger agent to produce debugging guidance, prompting the coder agent to refine the policy accordingly.

\begin{algorithm}[tb]
\caption{Iterative code generation}
\label{alg:coder}
\textbf{Input}: $message$, $trial_{max}$\\
\textbf{Output}: $code$
\begin{algorithmic}[1] 
\STATE $trial \leftarrow 0$
\STATE $code$ $\leftarrow$ Coder($message$)
\STATE $error$ $\leftarrow$ run($code$)
\WHILE{$error$ is not None}
\IF {$trial > trial_{max}$}
\STATE \textbf{return} "Failed to generate code within $trial_{max}$ "
\ENDIF
\STATE $guidance$ $\leftarrow$ Debugger($code$, $error$)
\STATE $code$ $\leftarrow$ Coder($guidance$)
\STATE $error \leftarrow$ run($code$)
\IF{$error$ is None}
    \STATE $error \leftarrow$ Evaluator($code$)
\ENDIF
\STATE $trial$ $\leftarrow$ $trial$ + 1
\ENDWHILE
\STATE \textbf{return} $code$
\end{algorithmic}
\end{algorithm}

\subsection{Reinforcement learning setup}
\label{RLsetup}
The RL problem can be formulated as a Markov Decision Process (MDP), defined by $(\mathcal{S}, \mathcal{A}, \mathcal{P}, r, \gamma)$, where 
$\mathcal{S}$ and $\mathcal{A}$ are the state space and action space, respectively. $\mathcal{P}: \mathcal{S} \times \mathcal{A} \times \mathcal{S} \rightarrow \mathbb{R}$ is the transition probability distribution, and $r: \mathcal{S} \times \mathcal{A} \rightarrow \mathbb{R}$ is the reward function, which is discounted by $\gamma \in [0,1)$ in temporal dimension. Within MDP, the goal of the RL agent is to learn a policy that maximizes the expected sum of discounted rewards.

In this work, we assume a policy network (actor) has already been trained to return action value from state variables, which is typical in policy-based and actor-critic methods. For demonstration purposes, we adopt the Soft Actor Critic (SAC) algorithm \citep{haarnoja2018soft} as our illustrative example. While we adopt SAC in our experiments, the proposed approach is readily generalizable to other policy-based or actor-critic algorithms.

\subsection{XRL tools}
\label{XRL tools}

\begin{figure}[t]
\centering
\includegraphics[width=0.7\columnwidth]{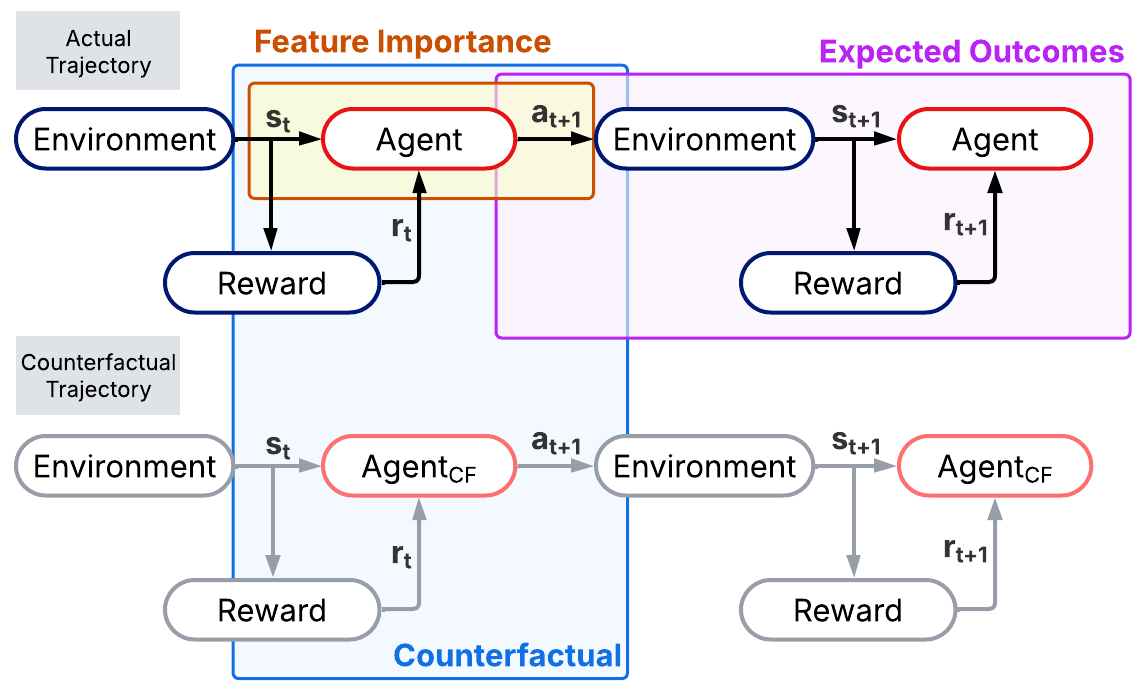} 
\caption{Explanatory scope of different XRL query types.}
\label{fig:2.XRLqueries}
\end{figure}

The central question in XRL is often framed as “Why did the agent execute this action?”. Since RL involves sequential decision-making with its goal of maximizing cumulative rewards, it offers multiple perspectives for interpreting agent behavior \citep{milani2024explainable}. As illustrated in Figure \ref{fig:2.XRLqueries}, we categorize the interpretation of RL agents into three perspectives: \textit{Feature Importance} (FI), \textit{Expected Outcomes} (EO), and \textit{Counterfactuals} (CF). We limit our scope to local explanations and leave \textit{policy-level} explanations for future work, to maintain the focus on specific state-action pairs.

To address each of these XRL tasks, we organize a set of XRL tools designed to interpret the local behavior of RL agents. These tools are automatically selected and called by LLM agents with contextually relevant arguments, enabling a more user-centric explanation.

\subsubsection{1) Feature Importance (FI)}
\label{Feature Importance}
approaches aim to identify which aspects of the current state most influence the agent’s specific action. In our framework, we adopt DeepSHAP (Deep SHapley Additive exPlanations) \citep{lundberg2017unified}, a post-hoc explainer tailored for deep neural networks, to estimate the contribution of each state variable to the agent’s action value. DeepSHAP can efficiently calculate feature attributions for the entire neural network by leveraging the chain rule, where detailed calculation procedure is elaborated in the Supplementary materials.

\subsubsection{2) Expected Outcome (EO)}
\label{Expected Outcome}
approaches aim to explain an agent’s behavior by analyzing anticipated future trajectories or rewards as a result of executing a particular action. EO explanation is especially valuable in domains where rewards are delayed, or instant inverse response could happen.

In this work, we decompose the agent’s Q-value, the expected discounted sum of future rewards. The Q-value can be decomposed in two complementary ways: (i) \textit{component-wise}, which splits the total reward into meaningful physical objectives, and (ii) \textit{temporally}, which breaks Q-value across time to identify when specific rewards are expected.

For component-wise decomposition, we follow the \textbf{code generation} scheme outlined in Algorithm \ref{alg:coder}. Given the Python code of the reward function used in RL training, the coder agent is instructed to first identify the semantic components of the reward and rewrite the function to return the decomposed sub-rewards for each component.

For the temporal dimension, we adopt a strategy inspired by \citet{van2018contrastive}, which used a simulation model to generate the most probable future trajectory conditioned on the current state and action. Although this approach assumes access to a simulation model, which may not always be available in real-world applications, we adopt this approach since it can offer post-hoc future trajectories without retraining the RL agent unlike other approaches \citep{ruggeri2025explainable, towers2024explaining}.

A resulting decomposition of Q-values is presented in Equation \ref{eq:Q decom}:

\begin{align}
Q^{\pi}(s,a) &= \mathbb{E} \left[ \sum_{k=1}^K \sum_{t=0}^{\infty} \gamma^t r_{t,k} | s_0 = s, a_0 = a, \pi \right]
\label{eq:Q decom}
\end{align}
where $k$ denotes component index, $K$ represents total number of components, and $\gamma$ is a discount factor.

\subsubsection{3) Counterfactuals (CF)}
\label{Counterfactuals}
explanations aim to answer contrastive questions such as ``What if?'' or ``Why not?'', highlighting why the agent selected the current action over plausible alternatives. The key limitation of the previous XRL studies on counterfactual explanation is that they cannot capture high-level demands beyond simple actions, which restrict their applicability \citep{gajcin2024redefining}.

To enhance the flexibility of counterfactual reasoning in RL practices, we introduce three novel types of counterfactual explanations.

\begin{itemize}

\item \textbf{Action-based Counterfactual (CF-A)} approach is conceptually similar to those used in previous XRL studies, where the counterfactual poses contrastive actions for a certain timestep (e.g., ``Why don't we take action $b$, instead of action $a$ at time $t$?'').

In our approach, we extend this concept by allowing counterfactual actions to be kept over a specified time interval, providing richer insights into how sustained action changes influence the system’s behavior.

\item \textbf{Behavior-based Counterfactual (CF-B)} explanations construct scenarios from qualitative descriptions about agent behavior. This type of counterfactual is particularly useful in control domains, where alternative scenarios are often expressed using terms like \textit{aggressive} or \textit{conservative} without specifying exact numeric details. While this resembles the tuning process of a PID (Proportional-Integral-Derivative) controller, there are no explicit parameters corresponding to such behaviors for RL agents. Therefore, it is necessary to interpret the nuance of the user’s query and translate it into appropriate numerical adjustments.

To achieve this, we introduce \textit{smoothing factor} $\alpha$ that adjusts the overall behavior of the action trajectory. For terms like \textit{aggressive} or \textit{conservative} trajectory, the idea of Polyak averaging \citep{polyak1992acceleration} is employed as shown in Equation \ref{eq:CF-B1}. It serves to modulate the sensitivity of a trajectory to temporal changes by adjusting the value of $\alpha$. For example, \textit{aggressive} trajectories, which respond to temporal changes, are assigned $\alpha > 1$, while \textit{conservative} trajectories, where the action values evolve gradually, are set to $0 < \alpha < 1$. Semantically similar terms (e.g., \textit{calm}, \textit{bold}, or \textit{steady}) are also given appropriate $\alpha$ values, owing to LLM's robustness in linguistic nuances. For \textit{opposite} behaviors, Equation \ref{eq:CF-B2} is used instead, as values may diverge if we directly use $\alpha < 0$ in Equation \ref{eq:CF-B1}.


\begin{align}
a_{t}^{new} &= a_{t-1}^{new} + \alpha (a_{t}^{prev} - a_{t-1}^{new})
\label{eq:CF-B1}
\\
a_{t}^{new} &= a_0^{prev} + \alpha (a_t^{prev} - a_0^{prev})
\label{eq:CF-B2}
\end{align}

\item \textbf{Policy-based Counterfactual (CF-P)} addresses a broader question of how a fundamentally different control strategy would affect future trajectories, rather than just a localized action deviation. For example, a user might want to explore whether a simple on-off controller—one that operates between only two discrete values conditioned on state variables— could serve as an alternative to a RL policy, and compare their outcomes.

To support such queries, We employ the \textbf{code generation} scheme Algorithm \ref{alg:coder} to generate a new executable policy code that faithfully reflects the user's intent. The generated policy is then executed in forward simulations to generate the rollout data.

\end{itemize}

In this work, we utilize generated counterfactual scenarios to get insights from contrasting the future state variables and rewards. Notably, those counterfactual scenarios can also be used in conjunction with EO approaches, by comparing the expected decomposed rewards of counterfactual trajectories with those of the original policy.

\section{Experimental results}

\subsection{Quadruple tank system}

\begin{figure}[t]
\centering
\includegraphics[width=0.5\columnwidth]{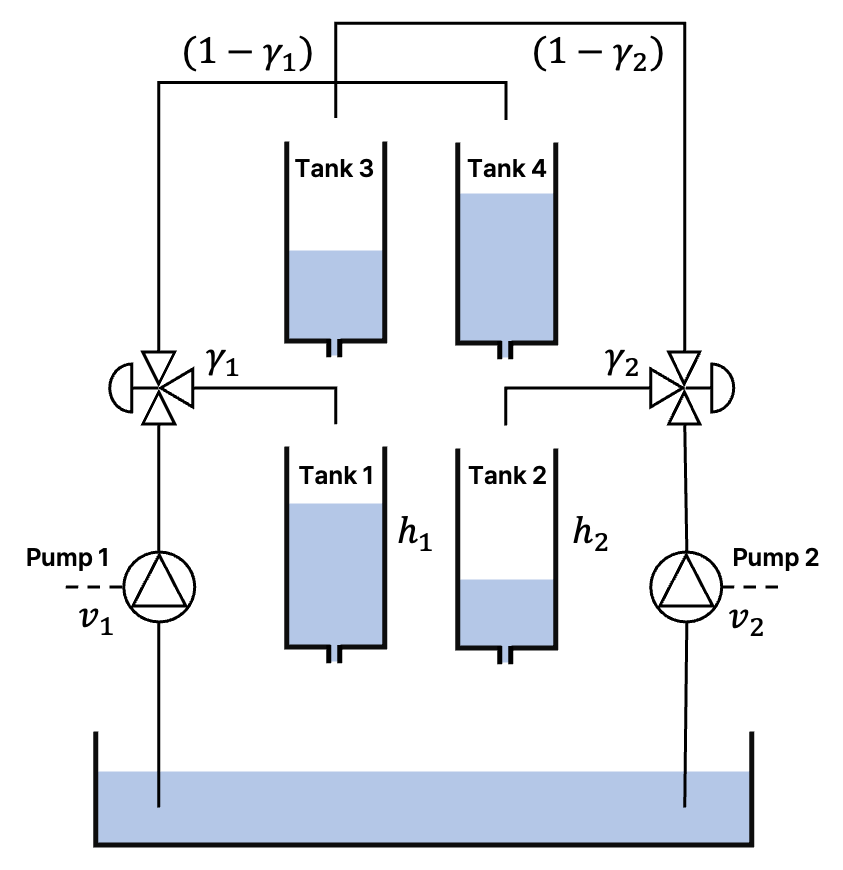}
\caption{Illustration of quadruple tank system benchmark}
\label{fig:quadruple}
\end{figure}

We evaluated our approach on the quadruple tank system, a well-known benchmark for RL-based control \citep{johansson2002quadruple}.  As illustrated in Figure \ref{fig:quadruple}, the system consists of four interconnected tanks, where water flows from two upper tanks to two lower tanks via gravity. The primary objective is to maintain specific water levels in the lower tanks ($h_1, h_2$) by tracking desired setpoints—predefined target values specified by the operators. The pumped water is distributed to different tanks, introducing strong coupling between state variables within the system. This coupling induces potential competition between multiple reward components, making the system a suitable benchmark for evaluating XRL approaches.

The action variables are the voltages applied to two pumps ($v_1, v_2$), which directly influence water flow. The state variables encompass the water levels in all four tanks, as well as the errors between the lower tank levels and their respective setpoints. The reward function is defined as the negative sum of the squared errors between the current $h_1, h_2$ values and their setpoints, where a minor penalty component is included to discourage abrupt changes in the action variables.

Although our approach is demonstrated in the context of process control, it can be readily generalized to other domains involving sequential decision making as long as a policy network of the agent is made available.


\subsection{Settings}
For training the RL models, we employed Stable-Baselines3 \citep{raffin2021stable}, while the quadruple tank benchmark was simulated using the PC-gym package \citep{bloor2024pc}. During training, the setpoints of the control objectives were periodically shifted to ensure that the learned policy could effectively track different values of setpoints. After training, we obtained a SAC agent capable of accurately regulating the setpoints. Details about environmental parameters and the training process of the SAC agent are reviewed in the supplementary material.
All experiments were conducted on a Windows machine with 32GB RAM, an Intel i9-12900K CPU and an NVIDIA GeForce RTX 3060 GPU.

In this study, our experiments were designed to address the following three key research questions:

\begin{itemize}
\item \textbf{RQ1}: Can \textit{TalkToAgent} accurately interpret user queries and call relevant XRL functions with correct arguments?

\item \textbf{RQ2}: Can \textit{TalkToAgent} reliably generate counterfactual scenarios that reflect user’s intention?

\item \textbf{RQ3}: Can \textit{TalkToAgent} correctly interpret the XRL results with domain-specific implications?
\end{itemize}

\subsection{Task classification accuracy}

\begin{table}[t]
\centering
\setlength{\tabcolsep}{1mm}


\begin{tabular}{lcc}
\toprule
Setting & few-shot & zero-shot\\
\midrule
gpt-4.1      & \bm{$97.5 \pm 0.9$} & $96.6 \pm 0.8$ \\
gpt-4o       & $96.6 \pm 0.3$ & $95.7 \pm 0.3$ \\
gpt-4.1-mini & $93.3 \pm 0.0$ & $91.8 \pm 0.9$ \\
gpt-4.1-nano & $88.7 \pm 1.1$ & $85.0 \pm 1.0$ \\
\bottomrule
\end{tabular}

\caption{Task classification accuracy (\%) across different GPT models in zero-shot and few-shot settings.}
\label{tab:coordinator}
\end{table}

To address RQ1, we first evaluated the coordinator’s ability in classifying user queries into the appropriate XRL function types. We generated a total of 90 example queries, comprising 20 queries for FI, EO CF-A, CF-B each, and 10 queries for CF-P type. We compared task classification accuracy of four different GPT variants (GPT-4.1, GPT-4o, GPT-4.1-mini, GPT-4.1-nano) under two prompt settings: \textit{zero-shot}, where only function descriptions are given, and \textit{few-shot}, where a few labeled examples are appended. The experiment was repeated 10 times independently for consistency.

The results in Table \ref{tab:coordinator} suggest that GPT-4.1 with few-shot prompting achieved the highest accuracy in task classification. Other models other than GPT-4.1-nano also effectively mapped user queries to the appropriate tasks, with accuracies higher than 90\%. Furthermore, providing only function descriptions proved sufficient for correctly identifying the query type, although with a slight reduction in performance. Notably, most inaccuracies are mainly observed in CF-B queries, such as ``keep the same action'', which could also have been handled by CF-A or CF-P solutions.

Moreover, the GPT-4.1 model achieved 100\% accuracy in extracting the correct function arguments, including mapping qualitative behavior terms to proper numerical $\alpha$ values in CF-B queries. The results ensure reliable selection of appropriate XRL tools in response to natural language queries.

\subsection{Counterfactual generation accuracy}

\begin{figure}[t]
\centering
\includegraphics[width=0.6\columnwidth]{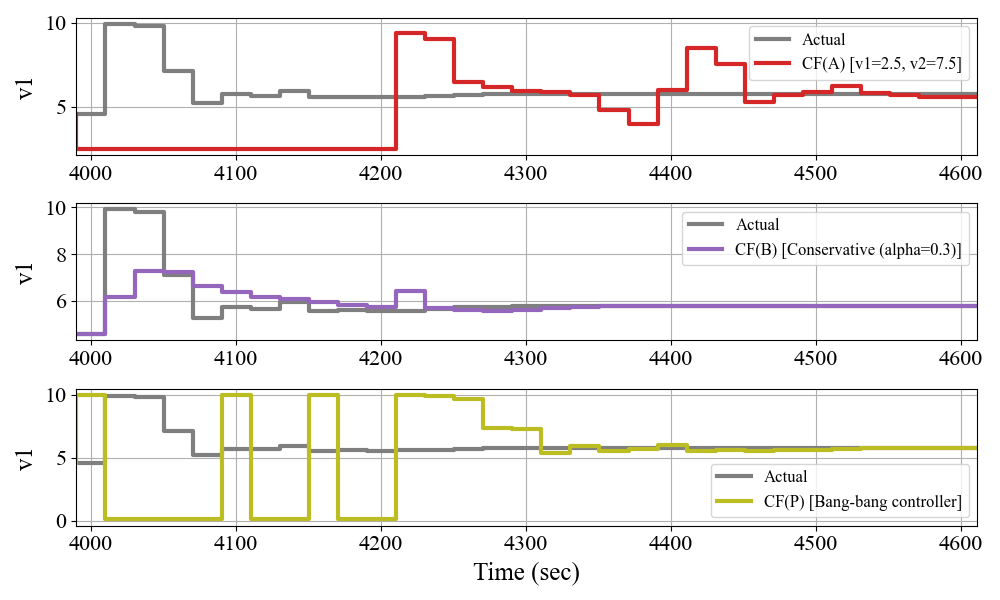}
\caption{Generated action trajectories for different counterfactual scenarios: (top) "What if we set the voltage of pump 1 to 2.5? (CF-A)"; (middle) "Why don't we get a more conservative behavior with $\alpha=0.3$? (CF-B)"; and (bottom) "Why don't we use an on-off controller instead? (CF-P)". All counterfactuals are applied from $t=4000$ to $4200$, after which control is reverted to the original RL policy.}
\label{fig:cf}
\end{figure}

\begin{figure}[t]
\centering
\includegraphics[width=0.7\columnwidth]{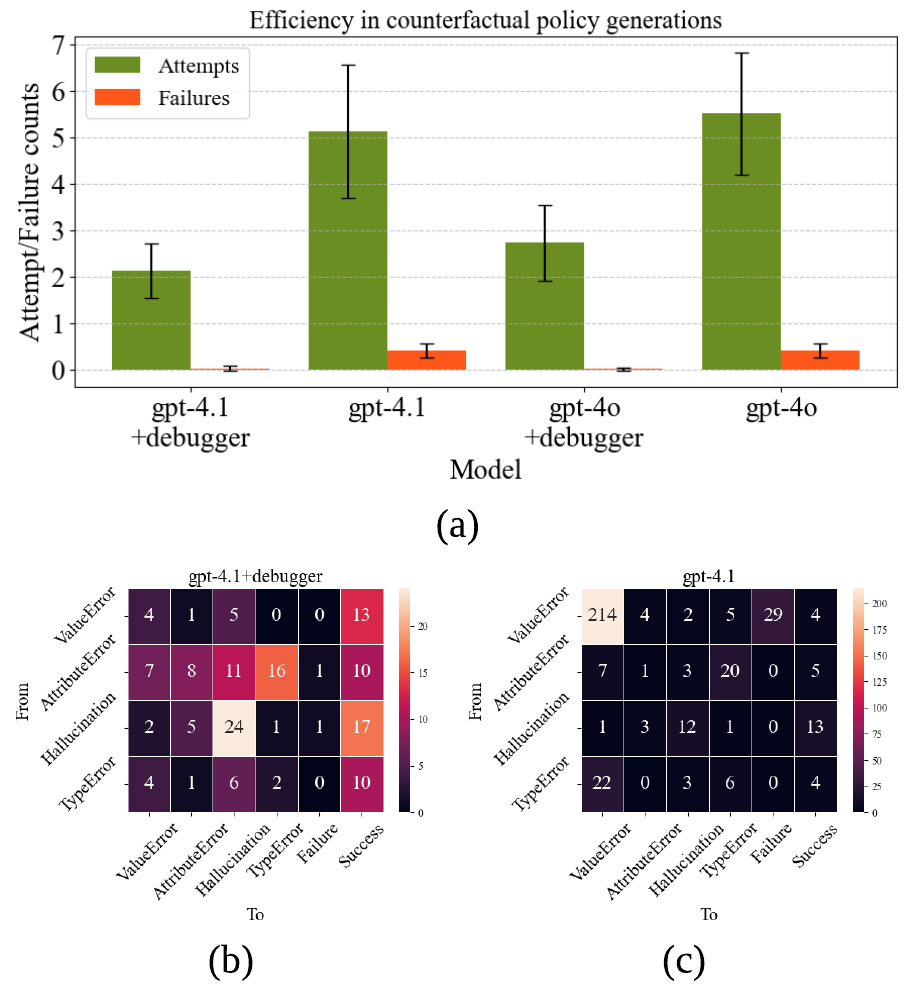}
\caption{(a) Average total iterations and failure counts for each LLM model. (b,c) Error transition matrices showing how errors evolve over iterative trials for the GPT-4.1 model—(b) with the debugger, and (c) without.}
\label{fig:cfgeneration}
\end{figure}

\begin{figure*}[t]
\centering
\includegraphics[width=0.95\textwidth]{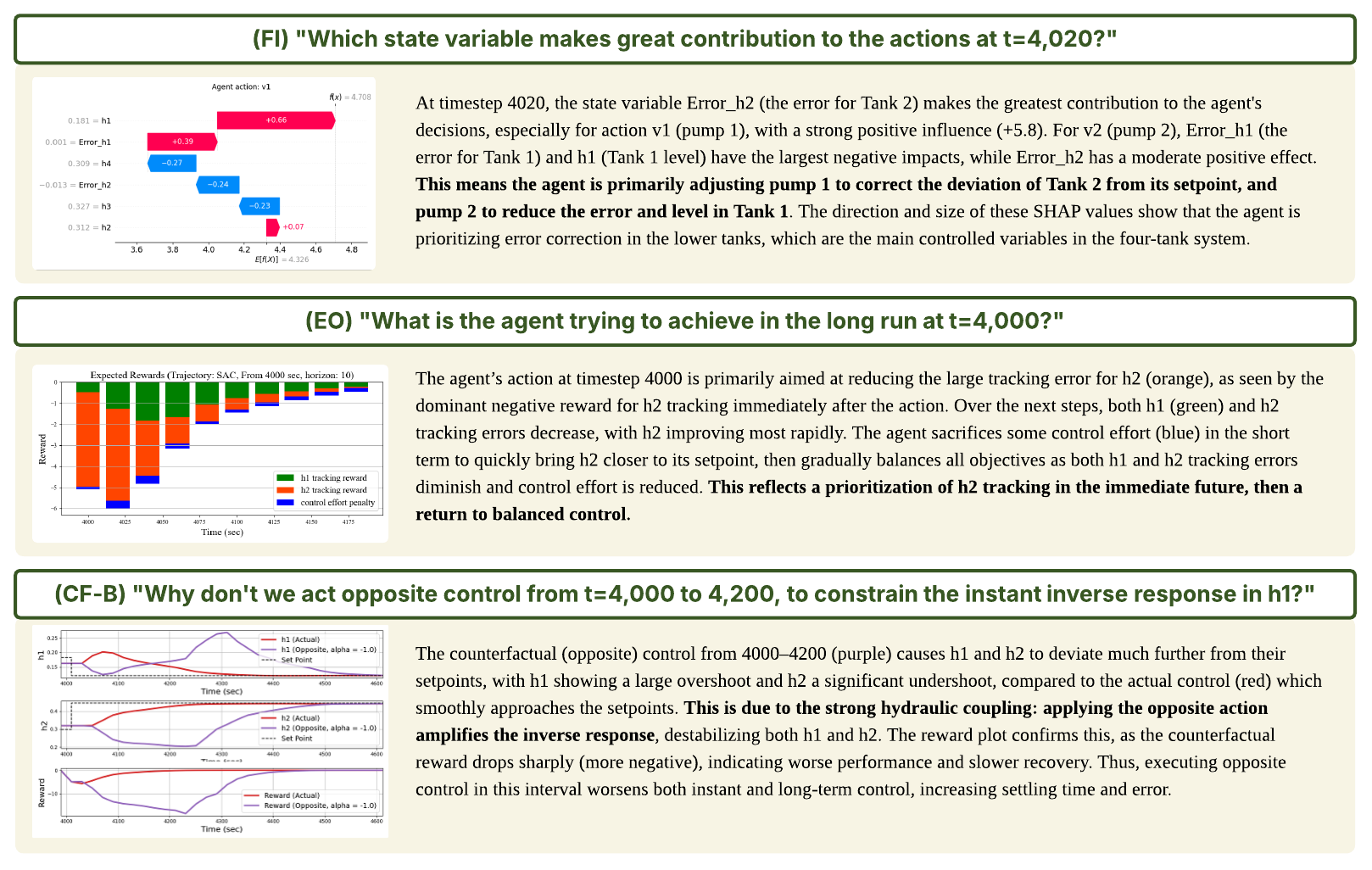} 
\caption{Illustrative results and explanations generated by \textit{TalkToAgent}, for three types of queries: (top) FI, (middle) EO, and (bottom) CF-B.}
\label{fig:EX3}
\end{figure*}

To address RQ2, we evaluated the fidelity of counterfactual trajectory generations. Figure \ref{fig:cf} presents the rollout action trajectories for the three counterfactual types compared with the actual trajectory.
In contrast to CF-A and CF-B queries where the intended trajectory is directly generated by the predefined functions, CF-P queries involve further code generation from LLM model, making it susceptible to generating hallucinated policies. This requires further validation to guarantee the fidelity of the resulting trajectories.

We tested 10 different CF-P queries by prompting \textit{TalkToAgent} to generate corresponding policies. The maximum number of attempts ($trial_{max}$ in Algorithm \ref{alg:coder}) was set to 10, and any inability to produce a valid policy within this limit was counted as a failure. To ensure consistency, the experiment was repeated across 7 independent trials. Two GPT models (GPT-4.1, GPT-4o) were tested, and an ablation study with the debugger agent was further studied. For each setting, we measured both the total number of attempts required and failure counts.

The ablation study in Figure \ref{fig:cfgeneration}a indicates that the use of debugger significantly reduce both the average number of attempts and total failures. To further understand how errors evolve during iterative refinement, we tracked all errors that the agent encountered during counterfactual policy generation. These errors were then classified by their type and content, and the results are illustrated as transition matrices in Figure \ref{fig:cfgeneration}b,c, which summarize the types of error transitions observed across all iterations of the \textit{coder-debugger} loop. The details of error types and their classification are in the supplementary material. We labeled an attempt as \textit{success} if the agent generated counterfactual policy code without any errors or hallucinations, and \textit{failure} if it failed to generate code after 10 trials. We counted all transitions between error types for 10 queries over all 7 iterations. The results show that without the debugger, the system often became stuck on recurring errors, especially on ValueError. In contrast, the debugger guided the agent toward a successful policy despite occasional hallucinations, which were also eventually resolved. This demonstrates the debugger’s effectiveness in providing meaningful guidance that helps the code converge to a valid solution.


\subsection{Qualitative evaluation of explanations}

To address RQ3, we qualitatively evaluated the multimodal explanations generated for FI, EO, and CF-B queries. For each query, both XRL plots and corresponding text explanations were generated, as illustrated in Figure \ref{fig:EX3}. Full explanations for all queries discussed in this paper are provided in the supplementary material.

First, for FI queries, the explanation revealed which state variables most strongly influenced each action variable. For example, as shown in Figure \ref{fig:EX3}, pump $v_1$ was primarily driven by the error term of $h_2$, while pump $v_2$ was most influenced by the error term of $h_1$.

For EO queries, the explanation highlighted how the agent prioritized rewards over time. We can notice that the agent initially focused on minimizing the $h_2$-tracking reward and later reduced both tracking rewards simultaneously.

Lastly, for CF-B queries, an \textit{opposite} control behavior to mitigate the inverse response of $h_1$ is compared with the original trajectory. The resulting state and reward trajectories revealed that while the inverse response in $h_1$ was alleviated instantly, but it amplified the inverse response in $h_2$, ultimately degrading the overall control performance. Importantly, this phenomenon was well explained in relation to the coupling between control variables.

Overall, these results demonstrate that the system was able to generate accurate and intuitive explanations for diverse natural language queries by combining XRL visualizations with domain-aware contexts.

\section{Conclusion}
In this work, we presented \textit{TalkToAgent}, a multi-agent LLM framework that bridges the gap between complex RL policies and domain experts with natural language interactions. Through experiments on the quadruple-tank system, we demonstrated its ability to accurately map user queries to suitable XRL tools, generate counterfactual explanations with reduced failure, and provide concise but accurate multimodal explanations.


A promising future direction is to extend our counterfactual methods to a result-oriented approach. While our current approach requires users to explicitly specify the counterfactual actions, behaviors, or policies, a goal-driven approach could infer relevant counterfactual actions or policies from a user's desired outcome. This would answer high-level questions such as “What would happen if actions are adjusted to concentrate solely on maximizing a single reward component?” without specifying low-level actions.

Furthermore, extending our approach to value-based networks and conducting case studies to various domains would be a promising future work.

\clearpage

\bibliographystyle{plainnat}
\bibliography{references}

\clearpage

\section*{S1. DeepSHAP calculation details for feature importance queries}
In our work, DeepSHAP is employed to address feature importance (FI) queries. The attribution of feature $y_k$ for function $h$ is calculated as follows:

\begin{align}
m_{xf}^j &= \frac{\phi_j(f,x)}{x_j - \mathbb{E}(x_j)}
\label{eq:multiplier}\\
m_{yh}^k &= \sum_{j=1}^nm_{yg_j}^k m_{xf} ^j
\label{eq:chain rule}\\
\phi_j(h,y) &= m_{yh}^k \left( y_j - \mathbb{E}(y_j) \right)
\label{eq:deepshapout}
\end{align}
where $m_{xf}^j$ denotes a multiplier for feature $x_j$ in function $f$, and $\phi_j(f,x)$ represents the corresponding Shapley value. Since Shapley values can be easily computed for simple blocks such as linear layers and activation functions, DeepSHAP can efficiently calculate feature attributions for the whole network by leveraging the chain rule. It first derives multipliers for each building block using Eq. \ref{eq:multiplier}, then recursively applies Eq. \ref{eq:chain rule} across all layers. The final Shapley value $\phi_j(h,y)$ for each feature is obtained via Eq. \ref{eq:deepshapout}.

\section*{S2. Experimental setting details}

\subsection*{S2.1. Parameter for quadruple tank system}

\begin{table}[h!]
\centering
\caption*{Table S1: Quadruple-tank environment parameters used for simulation}
\label{tab:four_tank_env}
\begin{tabular}{ll}
\toprule
\textbf{Parameter} & \textbf{Value / Description} \\
\midrule
Total simulation time ($T$) & 8000 seconds \\
Number of steps ($n_\text{steps}$) & 400 \\
Time step ($\Delta t$) & 20 seconds \\
Action variables & $v_1$, $v_2$ \\
State variables & $h_1$, $h_2$, $h_3$, $h_4$, $error_{h1}$, $error_{h2}$ \\
Action space (low) & $[0.1, 0.1]$ \\
Action space (high) & $[10, 10]$ \\
State space (low) & $[0, 0, 0, 0, -0.6, -0.6]$ \\
State space (high) & $[0.6, 0.6, 0.6, 0.6, 0.6, 0.6]$ \\
Setpoint range & Uniformly sampled in [0.1, 0.5] every 40 steps \\
Initial state & $[0.141, 0.112, 0.072, 0.42, 0.0, 0.0]$ \\
\bottomrule
\end{tabular}
\end{table}

Table S1 summarizes the simulation parameters used in the quadruple-tank control environment, including observation and action spaces, initial conditions, and setpoint generation scheme.

The reward at each timestep $t$ is defined as the negative sum of squared normalized tracking errors for the two controlled tank levels ($h_1$, $h_2$) and a control effort penalty term. Formally, the reward is given by:
\[
r_t = - \left[
100 \left( \hat{x}_{h_1,t} - \hat{s}_{h_1,t} \right)^2 +
100 \left( \hat{x}_{h_2,t} - \hat{s}_{h_2,t} \right)^2 +
\left\| \hat{u}_t - \hat{u}_{t-1} \right\|^2
\right]
\]
where $\hat{x}_{h_k,t}$, $\hat{s}_{h_k,t}$ are the normalized heights and their setpoints of tank $k \in \{1, 2\}$, respectively, and $\hat{u}_t$ and $\hat{u}_{t-1}$ are the normalized control input vectors at time $t$ and $t{-}1$, respectively.

\subsection*{S2.2. RL model training setup}
Table S2 presents the hyperparameters used for training the reinforcement learning (RL) agent.

\begin{table}[H]
\centering
\caption*{Table S2: RL training settings}
\label{tab:rl_training_settings}
\begin{tabular}{ll}
\toprule
\textbf{Parameter} & \textbf{Value / Description} \\
\midrule
RL algorithm & SAC (Soft Actor-Critic) \\
Total training steps ($n_\text{train}$) & $10^6$ \\
Learning rate & 0.001 \\
Discount factor ($\gamma$) & 0.9 \\
\bottomrule
\end{tabular}
\end{table}

\begin{figure*}[h!]
\centering
\includegraphics[width=0.8\textwidth]{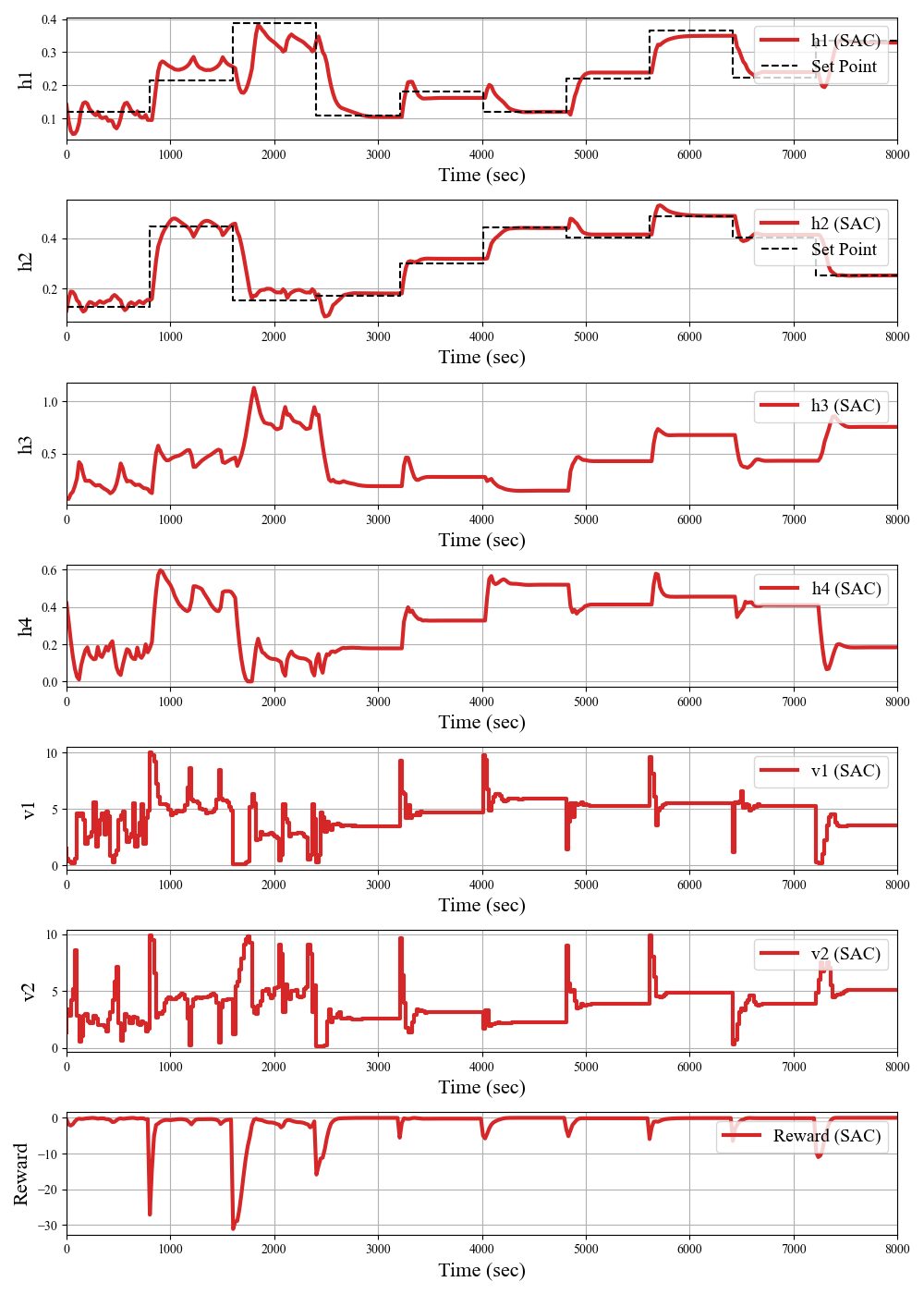} 
\caption*{Figure S1: Rollout results of the trained SAC agent in the quadruple-tank system, showing tank water levels ($h_1$, $h_2$, $h_3$, $h_4$), pump voltages ($v_1$, $v_2$), and reward.}
\label{fig:rollout}
\end{figure*}

Figure S1 presents the rollout results obtained by the SAC agent trained with the above hyperparameters. The figure illustrates the water heights of the four tanks, the control inputs (pump voltages), and the corresponding reward values. As shown, the trained SAC agent exhibits slight oscillatory behavior during the initial phase, but after about 3,000 seconds, it successfully tracks the setpoints for $h_1$ and $h_2$. The reward remains close to zero for most of the rollout, except for instant drops when setpoints change abruptly.

Notably, inverse responses can be observed at certain time points, where the control action initially moves in the opposite direction of the setpoint change. For example, this behavior appears in $h_1$ around 4,000 seconds and in $h_2$ around 5,000 seconds. Such phenomena arise due to the coupling between action and state variables in the system dynamics. To highlight and analyze these behaviors, these instances are specifically selected for constructing example queries in this work.

\clearpage

\section*{S3. Prompts used for LLM agents}
To enable each of the five agents in the TalkToAgent framework to perform their designated roles effectively, we design specialized prompts tailored to the functionality and context of each agent. These prompts provide explicit instructions, structural constraints, and contextual descriptions, ensuring that each agent behaves consistently and accurately within the overall process control task. The following sections describe the detailed prompt formulations used for each agent.

\subsection*{S3.1. Coordinator}
For the \textbf{Coordinator} agent, we incorporate the system description and environmental parameters to the prompt, to enable context-aware selection of appropriate XRL tools.

\begin{tcolorbox}[title=Coordinator prompt, colback=gray!5, colframe=gray!80!black, fonttitle=\bfseries]
\small
Your task is to choose the next function to work on the problem based on the given function tools and user queries.

The brief explanation of control system is given below: \texttt{\{system\_description\}}

-----

Furthermore, the environment parameters are given below: \texttt{\{env\_params\}}

-----

Here are a few points that you have to consider while calling a function:
\begin{itemize}
  \item When calling a function with \texttt{action} argument, make sure the action is within \texttt{env\_params["actions"]}. Otherwise raise an error.
  \item When queried for a certain time interval, make sure to use the queried time itself when calling the function, without dividing by \texttt{delta\_t} parameter.
  \item Also, don't scale neither state or action value, since it will be scaled at the subsequent functions.
\end{itemize}
\end{tcolorbox}

\subsection*{S3.2. Explainer}
For the \textbf{Explainer} agent, we augment the prompt with both the system description and environmental parameters to ensure that XRL results can be interpreted within the correct system context. To enhance interpretability for end users, the prompt also explicitly includes descriptions of the specific XRL tool used (\texttt{\{fn\_description\}}) and guidance on how to read the generated visualizations (\texttt{\{figure\_description\}}). The contents of these placeholders vary depending on which XRL tool is selected. Additionally, we introduce the \texttt{max\_token} parameter into the prompt, allowing users to flexibly adjust the level of detail by increasing or decreasing the token limit depending on their needs.

\begin{tcolorbox}[title=Explainer prompt, colback=gray!5, colframe=gray!80!black, fonttitle=\bfseries]
\small
You're an expert in both explainable reinforcement learning (XRL) and process control. Your role is to explain the user queries based on XRL results and related figures triggered by XRL functions.

\vspace{0.5em}
\textbf{User query:} \texttt{\{user\_query\}}

\vspace{0.5em}
Below are the name of the XRL function triggered and it's description:
\begin{itemize}
  \item \textbf{Function name:} \texttt{\{fn\_name\}}
  \item \textbf{Function description:} \texttt{\{fn\_description\}}
\end{itemize}

\vspace{0.5em}
- Also, for more clear explanation, the description of the system and its environment parameters are given as below:
\begin{itemize}
  \item \textbf{System description:} \texttt{\{system\_description\}}
  \item \textbf{Environment parameters:} \texttt{\{env\_params\}}
\end{itemize}

\vspace{0.5em}
- If XRL visualization are available, briefly explain how to interpret the all given visualization results.
\begin{itemize}
  \item \textbf{Figure description:} \texttt{\{figure\_description\}}
\end{itemize}

\vspace{0.5em}
\textbf{Important Notes:}
\begin{itemize}
  \item If there are multiple agent actions to be explained, you will get sets of the plots. Make sure to interpret them individually.
  \item IMPORTANT! Make sure to relate the XRL results to input-output relationship within the system, based on the given system description.
  \item The explanation output must be concise and short enough (below \texttt{\{max\_tokens\}} tokens), because users may be distracted by too much information.
  \item Try to concentrate on providing only the explanation results, not on additional importance of the explanation.
\end{itemize}
\end{tcolorbox}

\subsection*{S3.3. Coder}
The \textbf{Coder} agent is designed with separate prompts tailored to the purpose and context of each task. Specifically, two main tasks are supported: counterfactual policy generation in response to policy-based counterfactual (CF-P) queries, and decomposed reward function generation for expected outcomes (EO) queries. To enable interactive debugging between the Coder and Debugger agents, the Coder also incorporates refining input messages that include debugging guidance passed from the Debugger agent via the \texttt{guidance} variable.

In addition, the generated code is instructed to include no additional comments or extraneous content. In order to address some occasional violations of this instruction we implement a separate \texttt{\_sanitize} function to remove any non-code elements from the output prior to execution. This efficiently addresses issues such as the inclusion of markdown syntax (e.g., triple quotes) in decomposed reward functions, or inline comments in \texttt{CF\_policy} classes.

\subsubsection*{S3.3.1. Counterfactual policy generation}
For counterfactual policy generation, the prompt explicitly provides a base structure for the \texttt{CF\_policy} class, ensuring consistency with the overall format of the original RL policy. To further allow the Coder to match the expected output shape, an example output (\texttt{output\_example}) from the original policy is also included in the prompt. The prompt also entails utility methods such as \texttt{self.env.\_scale\_U(u)}, enabling the agent to use them when needed.

In our work, the generated policy is limited to simple, rule-based policies that can be encoded with basic logic. To enforce this constraint, the prompt instructs the agent to raise an error when queries request more advanced control methods such as MPC(Model Predictive Control) or PID(Proportional-Integral-Derivative) controllers. Extending the framework to support such methods is a promising direction for future work.

\begin{tcolorbox}[title=Coder prompt – Counterfactual policy generation, colback=gray!5, colframe=gray!80!black, fonttitle=\bfseries]
\small
You are a coding expert that generates rule-based control logic, based on user queries.  
Your job is to write a code for the following Python class structure, named \texttt{CF\_policy}:

\vspace{0.5em}
\texttt{%
======================== \\
import numpy as np \\
np.random.seed(21) \\
\\
class CF\_policy(): \\
\quad def \_\_init\_\_(self, env, original\_policy): \\
\quad\quad self.env = env \\
\quad\quad self.original\_policy = original\_policy \\
\\
\quad def predict(self, state, deterministic=True): \\
\quad\quad \# INSERT YOUR RULE-BASED LOGIC HERE \\
\quad\quad return action \\
========================
}

\vspace{1em}
Please consider the following points when writing the \texttt{predict} method:
\begin{itemize}
  \item If the instruction requires you to modify the original policy, feel free to use the \texttt{self.original\_policy.predict(state)} method.
  \item The output of the \texttt{predict} method (i.e., the action) should be within the range \texttt{[-1, 1]}, as it will be used by an external function that expects scaled values.  
        You can scale the action values by using the method: \texttt{self.env.\_scale\_U(u)}, if needed.
  \item The input \texttt{state} is also scaled. Ensure that your if-then logic works with scaled variables.  
        To scale raw state values, you may use: \texttt{self.env.\_scale\_X(x)}.
  \item The input for the \texttt{predict} method (\texttt{state}) is the same shape as the initial state \texttt{x0}.
  \item The output for the \texttt{predict} method (\texttt{action}) should match the shape of the output from the original policy.  
        Example output: \texttt{\{output\_example\}}
  \item If your code requires any additional Python modules, make sure to import them at the beginning of your code.
  \item Only return the code of the \texttt{CF\_policy} class, \textbf{without any additional comments}.
  \item If the user requested controllers other than rule-based ones (e.g., MPC, PID), trigger the \texttt{raise\_error} tool.
\end{itemize}

\vspace{0.5em}
For accurate policy generation, here are some descriptions of the control system:  
\texttt{\{system\_description\}}

\vspace{0.5em}
Also, environment parameters used in process control:  
\texttt{\{env\_params\}}

\vspace{0.5em}
You will get a great reward if you correctly generate the counterfactual policy function!
\end{tcolorbox}

In the refinement stage, the input message incorporates both the error (or hallucination Exception raised by the Evaluator agent) encountered during policy execution and the corresponding debugging guidance generated by the Debugger agent. This information is provided to the Coder agent to support accurate correction of the previously generated policy. Importantly, the same constraints enforced during the initial generation—such as returning only the \texttt{CF\_policy} class without any additional content—are emphasized in the refining input message to ensure consistency in output format.

\begin{tcolorbox}[title=Coder input – Refine the counterfactual policy with guidance, colback=gray!5, colframe=gray!80!black, fonttitle=\bfseries]
\small
You previously generated the following code for a counterfactual policy:  
\texttt{\{self.prev\_codes[-1]\}}

\vspace{0.5em}
However, the following error occurred during simulation:  
\texttt{\{error\_message\}}

\vspace{0.5em}
In order to debug this error, our Debugger agent suggested the following guidance:  
\texttt{\{guidance\}}

\vspace{0.5em}
Please revise the code to fix the error. Only return the corrected \texttt{CF\_policy} class.  
Also, you must still follow the original instructions from the initial prompt when modifying the code.
\end{tcolorbox}

\subsubsection*{S3.3.2. Decomposed reward generation}

For decomposed reward generation, the prompt instructs the Coder agent to first interpret a given reward function and decompose the original scalar reward into multiple component-wise rewards. The agent is expected not only to generate the corresponding Python function that returns a tuple of individual rewards, but also to assign meaningful labels to each component based on their control objectives.

To facilitate semantic interpretation, the prompt explicitly requires the agent to output both the code and the list of concise component names, separated by a \texttt{\textbackslash n---\textbackslash n} delimiter. This formatting allows the two outputs to be programmatically parsed and used separately. Additionally, the system description is included in the prompt to help the agent generate physically and contextually appropriate component names.

\begin{tcolorbox}[title=Coder prompt – Decomposed reward generation, colback=gray!5, colframe=gray!80!black, fonttitle=\bfseries]
\small
Your job is to decompose reward function into multiple components.  
You will get a Python code of the reward function used to train the RL controller agent,  
and your job is to return its corresponding decomposed reward function.

\vspace{0.5em}
Here are some requirements help you decompose the reward:
\begin{enumerate}
    \item While the original reward function gives scalar reward, the decomposed reward should be in \texttt{tuple} format,  
    which contains each component reward.
    
    \item When returning answer, please only return the following two outputs:
    \begin{itemize}
        \item The resulting Python function code. It would be better if necessary Python packages are imported.  
        Remove unnecessary strings like \texttt{'''}, or \texttt{''' python}.
        
        \item List of concise names of each control objective component.  
        The format should be a Python list, so that we can directly translate it into a list object using the \texttt{ast} module.
    \end{itemize}
    These two outputs should be separated by the separating signal \texttt{\textbackslash n---\textbackslash n}.
    
    \item You will also be given a brief description of the system.  
    Please follow the description to appropriately decompose the reward.
    
    \item The function's name should be in the form of \texttt{(original\_function\_name)\_decomposed}.
\end{enumerate}

\vspace{0.5em}
Here are the description of the current system and corresponding environment parameters that may help you decompose the reward function:

\textbf{System description:} \texttt{\{system\_description\}}

\vspace{0.5em}
\textbf{Environment parameters:} \texttt{\{env\_params\}}

\vspace{0.5em}
You will get a great reward if you correctly decompose the reward!
\end{tcolorbox}

The refinement process for decomposed reward functions closely mirrors that of policy generation.

\begin{tcolorbox}[title=Coder input – Refine the decomposed reward with guidance, colback=gray!5, colframe=gray!80!black, fonttitle=\bfseries]
\small
You previously generated the following code for a decomposed reward function:  
\texttt{\{self.prev\_codes[-1]\}}

\vspace{0.5em}
However, the following error occurred during simulation:  
\texttt{\{error\_message\}}

\vspace{0.5em}
In order to debug this error, our Debugger agent suggested the following below:  
\texttt{\{guidance\}}

\vspace{0.5em}
Please revise the code to fix the error. Only return the corrected \texttt{(original function name)\_decomposed} function.  
Also, you still have to follow the instructions from the initial prompt when modifying the code.
\end{tcolorbox}

\subsection*{S3.4. Evaluator}

The \textbf{Evaluator} agent is instructed to assess whether the generated counterfactual policy faithfully reflects the user’s intention. If not, the agent is required to raise an error. As with other agents, the prompt includes the system description and environmental parameters to support context-aware evaluation.

\begin{tcolorbox}[title=Evaluator prompt, colback=gray!5, colframe=gray!80!black, fonttitle=\bfseries]
\small
You are an evaluator that determines whether the Coder agent has correctly implemented the counterfactual policy.  
If you believe the resulting trajectory faithfully follows the intention of the user message, you may confirm the policy.  
Otherwise, you must raise an error by calling the \texttt{raise\_error} tool.

\vspace{0.5em}
For accurate evaluation of the trajectory, here are some descriptions of the control system:  
\texttt{\{system\_description\}}

\vspace{0.5em}
Also, environment parameters used in process control:  
\texttt{\{env\_params\}}

\vspace{0.5em}
You will get a great reward if you correctly filter out any incorrectly generated counterfactual policies!
\end{tcolorbox}

\subsection*{S3.5. Debugger}
The \textbf{Debugger} agent generates a guidance message for the Coder agent, suggesting how to fix the current code based on the encountered error or hallucination exception. This guidance is informed by the provided system description and environmental parameters.

\begin{tcolorbox}[title=Debugger prompt, colback=gray!5, colframe=gray!80!black, fonttitle=\bfseries]
\small
You are a coding expert that provides proper guidance to the Coder agent  
to help refine the generated code and resolve the encountered error.

\vspace{0.5em}
For accurate debugging, here are some descriptions of the control system:  
\texttt{\{system\_description\}}

\vspace{0.5em}
Also, environment parameters used in process control:  
\texttt{\{env\_params\}}

\vspace{0.5em}
You will get a great reward if you correctly guide the Coder agent to refine the policy accurately!
\end{tcolorbox}

\section*{S4. Error categorization in counterfactual policy generation}
To classify the errors raised during counterfactual policy generation, we first conducted clustering on collected error messages. Each error string was embedded using OpenAI’s \texttt{text-embedding-3-small} model to obtain vector representations. We then applied K-Means clustering to group the errors into 6 clusters. The result, visualized in Figure~S2, was obtained by reducing the embedding dimension to two using Principal Component Analysis (PCA) method.

\begin{figure*}[h]
\centering
\includegraphics[width=0.5\textwidth]{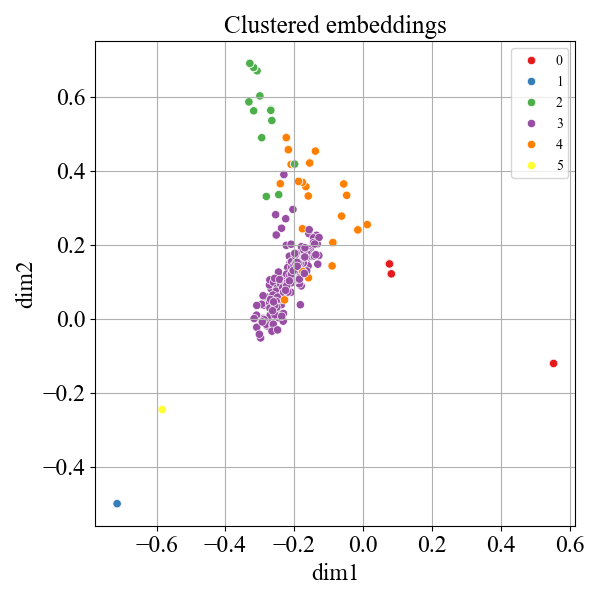} 
\caption*{Figure S2: Clustering results of text embeddings of error messages collected over 7 iterations of counterfactual policy generation experiments.}
\label{fig:cluster}
\end{figure*}

After clustering, we manually examined the error messages within each cluster to assign meaningful error categories, which are summarized in Table S3. We can observe that the error messages are generally clustered well according to their semantic type. Based on these categories, we analyze how errors evolved across iterations of the counterfactual policy generation process and constructed a transition matrix as discussed in the main text.


\begin{table}[H]
\centering
\caption*{Table S3: Representative error messages grouped by error category}
\label{tab:error_messages}
\begin{tabular}{>{\centering\arraybackslash}m{3cm} p{11cm}}
\toprule
\textbf{Error Category} & \textbf{Representative Error Messages} \\
\midrule

ValueError &
\tabitem setting an array element with a sequence. The requested array has an inhomogeneous shape after 1 dimensions. \\
& \tabitem operands could not be broadcast together with shapes (4,) (6,) \\
& \tabitem too many indices for array: array is 1-dimensional, but 2 were indexed \\
\midrule

AttributeError &
\tabitem 'make\_env' object has no attribute 'o\_space' \\
& \tabitem 'make\_env' object has no attribute 'unscale\_X' \\
& \tabitem make\_env.step() missing 1 required positional argument: 'action' \\
\midrule

Hallucination &
\tabitem The trajectory does not faithfully follow the user's intention because the action v1 was not consistently set to the maximum value whenever h1 $<$ 0.2 during the specified timestep range (4000 to 4400). \\
& \tabitem The trajectory does not implement the hybrid policy as specified. When h4 $>$ 0.7, v2 should be set to 1.5, but the provided control actions do not reflect this. \\
\midrule

TypeError &
\tabitem can only concatenate tuple (not "int") to tuple \\
& \tabitem unsupported operand type(s) for +: 'NoneType' and 'int' \\
& \tabitem float() argument must be a string or a real number, not 'NoneType' \\
\midrule

Failure &
\tabitem Failed after multiple attempts. \\
\midrule

Success &
\tabitem Code successfully generated. Rollout complete. \\
\bottomrule
\end{tabular}
\end{table}

\section*{S5. Multimodal explanation results for all queries}
We provide representative user queries and corresponding explanations for FI, EO, CF-A, CF-B, and CF-P tasks. In this work, a \texttt{max\_token} value of 200 was used in this work. Higher value of \texttt{max\_token} would allow for more detailed responses. All examples were generated using OpenAI’s GPT-4.1 model.


Note that the format of the responses may vary across runs, as no fixed output schema is enforced. Additionally, due to the inherent stochasticity of large language models, responses may not be fully deterministic even when using the same seed and temperature settings. Nevertheless, the XRL results themselves are fully reproducible, as they are generated by predefined functions.

\subsection*{S5.1. Feature Importance (FI)}
\textbf{Query:} Which state variable makes great contribution to the agent's decisions at t=4020?

\begin{figure*}[h!]
\centering
\includegraphics[width=1.0\textwidth]{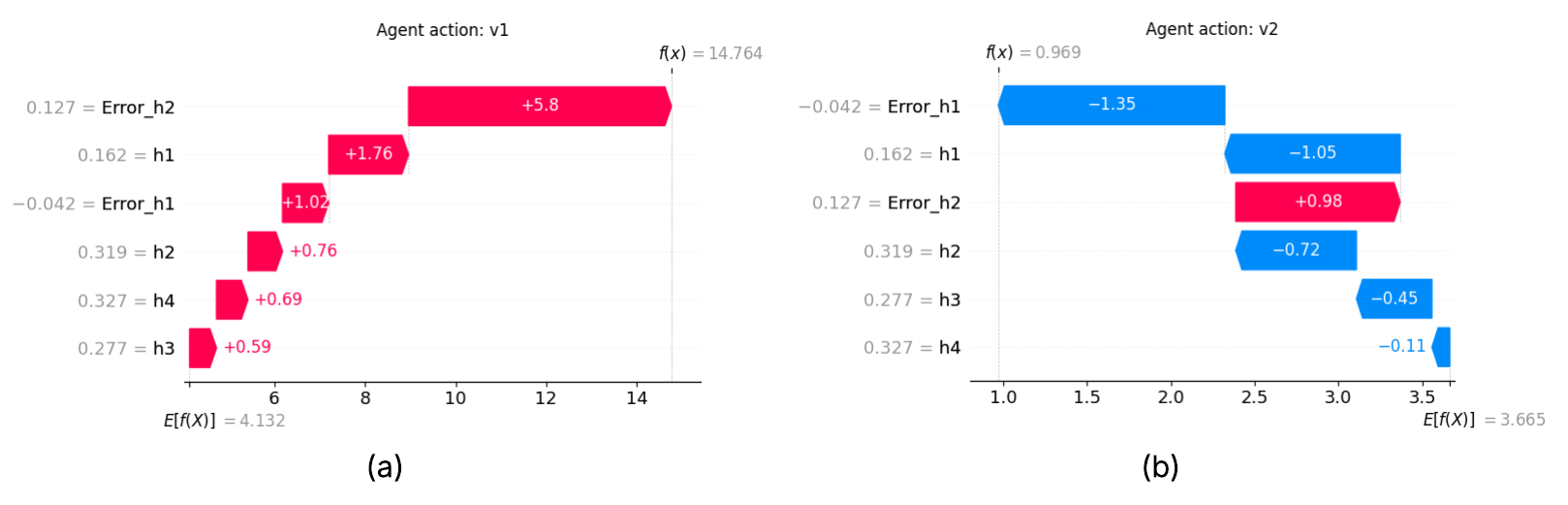} 
\caption*{Figure S3: SHAP results for Feature Importance (FI) query}
\label{fig:firesult}
\end{figure*}

\textbf{Answer:} At t=4020, the state variable **Error\_h2** (the error in Tank 2 level) makes the greatest contribution to the agent’s decision for action **v1**, with a strong positive effect (+5.8). For action **v2**, **Error\_h1** (the error in Tank 1 level) has the largest (negative) impact (-1.35). This means the agent is primarily adjusting v1 to correct Tank 2’s deviation from setpoint, and v2 to correct Tank 1’s deviation. The magnitude and direction of these contributions reflect the agent’s focus on minimizing the most significant errors in the lower tanks, which are the main control targets in the four-tank system. (Note that the XRL results and their corresponding explanations differ slightly from those in Figure 6, due to error corrections in the SHAP implementation within the XRL tools. This correction does not affect the main conclusions of the paper, though.)

\subsection*{S5.2. Expected Outcomes (EO)}
\textbf{Query:} What is the agent trying to achieve in the long run at t=4000?

\textbf{Answer:} The agent’s action at timestep 4000 is primarily aimed at reducing the large tracking error for h2 (orange), as seen by the dominant negative reward for h2 tracking immediately after the action. Over the next steps, both h1 (green) and h2 tracking errors decrease, with h2 improving most rapidly. The agent sacrifices some control effort (blue) in the short term to quickly bring h2 closer to its setpoint, then gradually balances all objectives as both h1 and h2 tracking errors diminish and control effort is reduced. This reflects a prioritization of h2 tracking in the immediate future, then a return to balanced control.

\begin{figure*}[h!]
\centering
\includegraphics[width=0.7\textwidth]{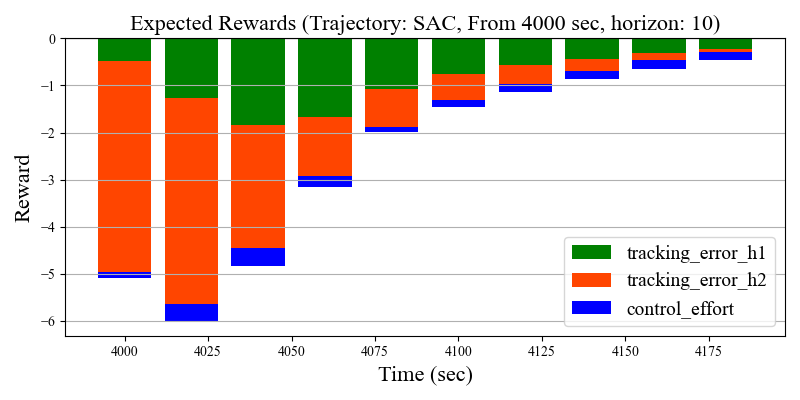} 
\caption*{Figure S4: Decomposed reward results for Expected Outcome (EO) query}
\label{fig:eoresult}
\end{figure*}

\subsection*{S5.3. Action-based Counterfactuals (CF-A)}
\textbf{Query:} Why don't we set the value of v1 action to 2.5 and v2 action to 7.5 from 4000 to 4200?

\begin{figure*}[h!]
\centering
\includegraphics[width=0.45\textwidth]{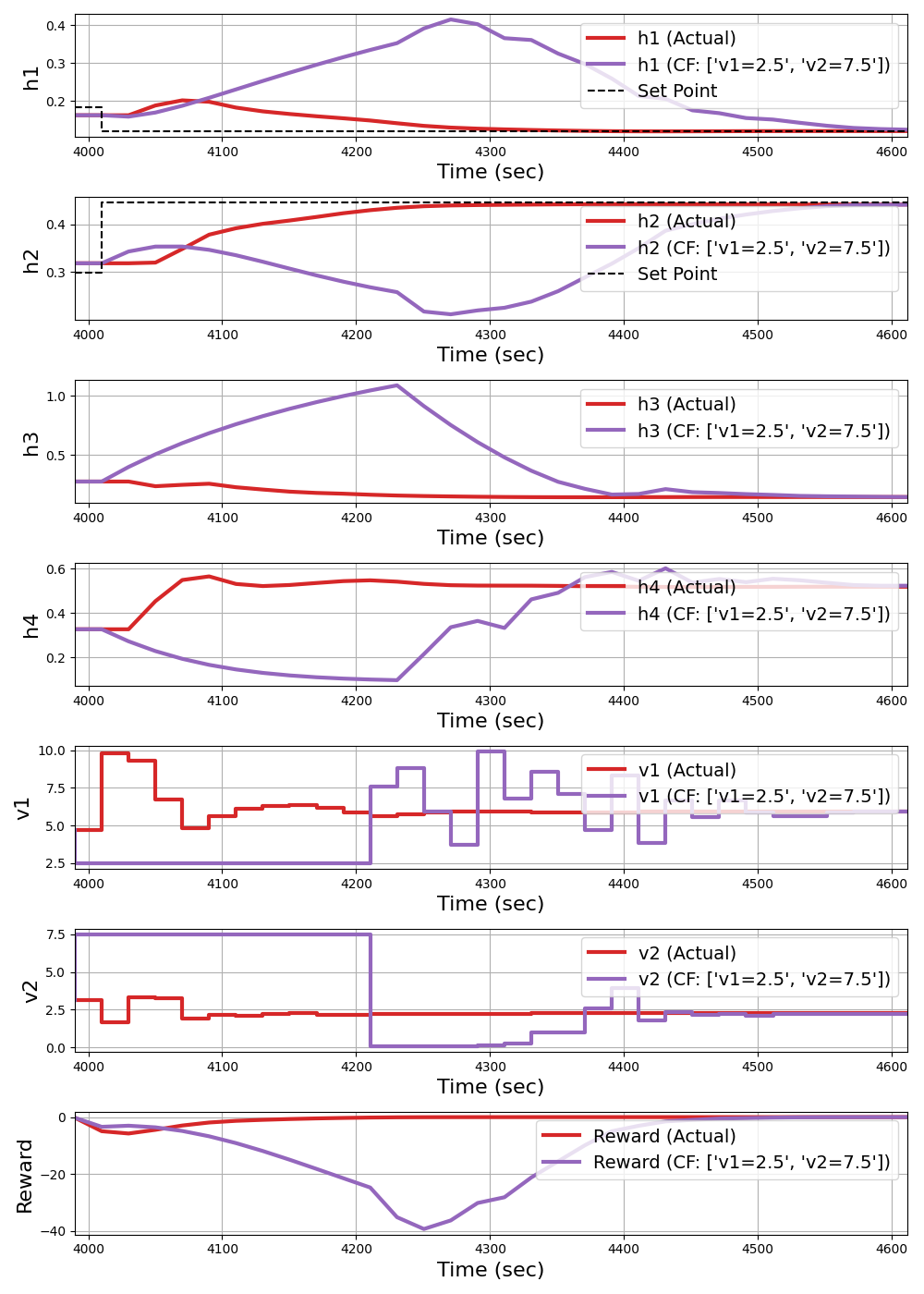} 
\caption*{Figure S5: Forward simulation results for Action-based Counterfactual (CF-A) query}
\label{fig:cfaresult}
\end{figure*}

\textbf{Answer:} Setting v1=2.5 and v2=7.5 from 4000 to 4200 (purple, CF) caused h1 and h2 to deviate significantly from their setpoints, with h1 overshooting and h2 undershooting, compared to the actual controller (red) which kept both close to target. This is reflected in the much lower reward (more negative) for the counterfactual, indicating worse control performance. The fixed actions led to large, slow oscillations in tank levels (especially h3, h4), and a longer settling time. Thus, the original controller’s adaptive actions are superior for maintaining stability and setpoint tracking in this multivariable system.

\subsection*{S5.4. Behavior-based Counterfactuals (CF-B)}
\textbf{Query:} Why don't we act opposite control from t=4000 to 4200, to constrain the instant inverse response in h1?

\begin{figure*}[h!]
\centering
\includegraphics[width=0.45\textwidth]{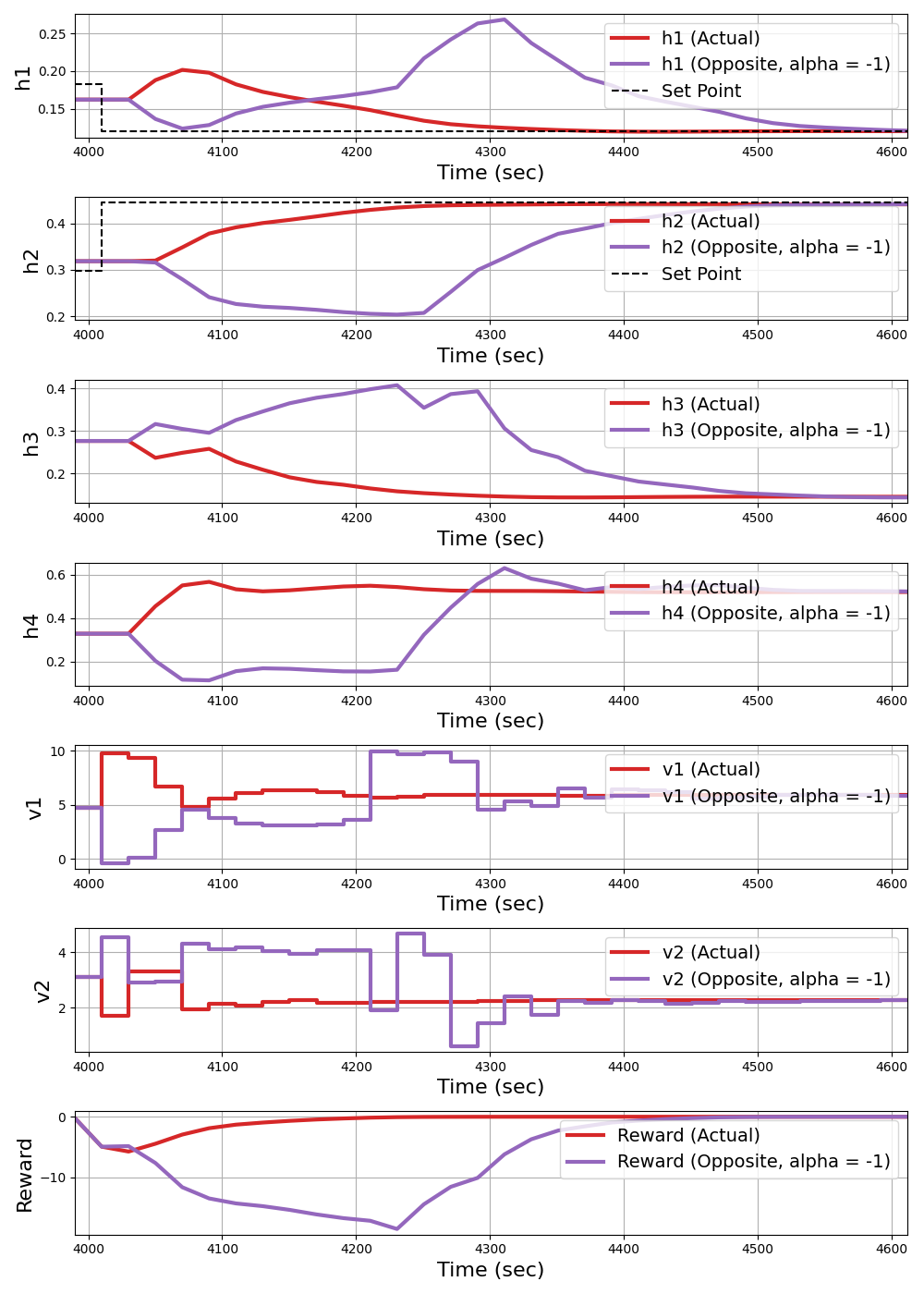} 
\caption*{Figure S6: Forward simulation results for Behavior-based Counterfactual (CF-B) query}
\label{fig:cfbresult}
\end{figure*}

\textbf{Answer:} The counterfactual (opposite) control from 4000–4200 (purple) causes h1 and h2 to deviate much further from their setpoints, with h1 showing a large overshoot and h2 a significant undershoot, compared to the actual control (red) which smoothly approaches the setpoints. This is due to the strong hydraulic coupling: applying the opposite action amplifies the inverse response, destabilizing both h1 and h2. The reward plot confirms this, as the counterfactual reward drops sharply (more negative), indicating worse performance and slower recovery. Thus, executing opposite control in this interval worsens both instant and long-term control, increasing settling time and error.

\subsection*{S5.5. Policy-based Counterfactuals (CF-P)}
\textbf{Query:} What would happen if we replaced the current RL policy with an on-off controller between 4000 and 4200 seconds, such that $v_1 = 8.0$ whenever the error of $h_1 < 0.0$, and $v_1 = 1.0$ otherwise; and similarly, $v_2 = 8.0$ whenever the error of $h_2 < 0.0$, and $v_2 = 1.0$ otherwise?"

\begin{figure*}[h!]
\centering
\includegraphics[width=0.45\textwidth]{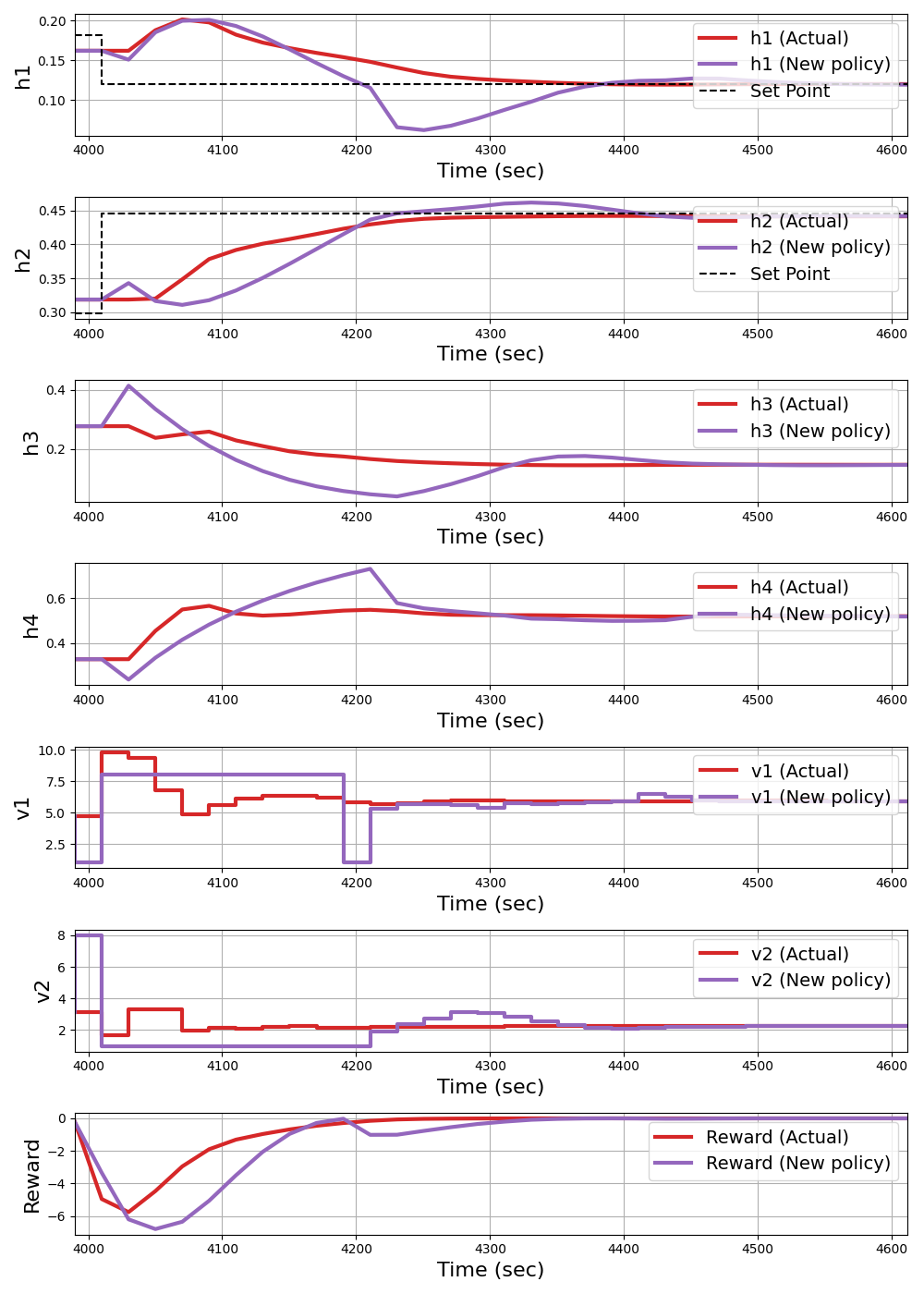} 
\caption*{Figure S7: Forward simulation results for Policy-based Counterfactual (CF-P) query}
\label{fig:cfpresult}
\end{figure*}

\textbf{Answer:} From 4000 to 4200 sec, the on-off controller (purple) causes h1 to drop sharply below its setpoint, while h2 rises more quickly and overshoots its setpoint compared to the RL policy (red). This is due to the aggressive switching of v1 and v2 between 1.0 and 8.0, leading to oscillatory and less stable tank levels. The reward trajectory for the on-off controller is consistently lower (worse) than the RL policy, indicating higher error and poorer control performance. After 4200 sec, both policies converge, but the on-off controller results in more overshoot and slower settling. Overall, the RL policy provides smoother, more stable control with less overshoot and better reward.

\end{document}